\pdfoutput=1

\documentclass[11pt]{article}

\usepackage[preprint]{acl}

\usepackage{times}
\usepackage{latexsym}

\usepackage[T1]{fontenc}

\usepackage[utf8]{inputenc}

\usepackage{microtype}

\usepackage{inconsolata}

\usepackage{graphicx}

\newcommand{\name}[0]{\textsc{ReliableEval}}
\newcommand{\numbenchmarks}[0]{three}
\newcommand{\nummodels}[0]{five}

\newcommand{\gpt}[0]{GPT-4o}
\newcommand{\llama}[0]{Llama-3.3-70B}

\newcommand{\grok}[0]{Grok-3}
\newcommand{\claude}[0]{Claude-3.7-Sonnet}
\newcommand{\deepseek}[0]{Deepseek-v3}

\newcommand{\mmlu}[0]{MMLU}
\newcommand{\gpqa}[0]{GPQA-Diamond}
\newcommand{\simple}[0]{SimpleQA}
\newcommand{\prompts}[0]{meaning-preserving prompt perturbations}

\usepackage{url}
\usepackage{amsfonts}
\usepackage{amsmath,amssymb}
\usepackage{amsthm}
\theoremstyle{definition}
\newtheorem{definition}{Definition}
\usepackage{bm}
\usepackage{booktabs}
\usepackage{subcaption}
\usepackage{stfloats}
\usepackage{placeins}
\usepackage[normalem]{ulem}

\title{\name{}: A Recipe for Stochastic LLM Evaluation\\via Method of Moments}

\author{
 \textbf{Gili Lior\textsuperscript{1}}\quad
 \textbf{Eliya Habba\textsuperscript{1}}\quad
 \textbf{Shahar Levy\textsuperscript{1}} \quad
 \textbf{Avi Caciularu\textsuperscript{2}}\quad
 \textbf{Gabriel Stanovsky\textsuperscript{1}}
\\
\\
 \textsuperscript{1}The Hebrew University of Jerusalem \quad
 \textsuperscript{2}Google Research
\\[0.5em]
\href{mailto:gili.lior@mail.huji.ac.il}{gili.lior@mail.huji.ac.il}
}

\begin{document}
\maketitle
\begin{abstract}
LLMs are highly sensitive to prompt phrasing, yet standard benchmarks typically report performance using a single prompt, raising concerns about the reliability of such evaluations. In this work, we argue for a stochastic method of moments evaluation over the space of \prompts. We introduce a formal definition of \emph{reliable evaluation} that accounts for prompt sensitivity, and suggest \name{} -- a method for estimating the number of prompt resamplings needed to obtain meaningful results. Using our framework, we stochastically evaluate five frontier LLMs and find that even top-performing models like \gpt{} and \claude{} exhibit substantial prompt sensitivity. Our approach is model-, task-, and metric-agnostic, offering a recipe for meaningful and robust LLM evaluation.\footnote{Code and data available at~\url{https://github.com/SLAB-NLP/Reliable-Eval}}
\end{abstract}

\section{Introduction}

\begin{figure}[t!]
    \centering
    \includegraphics[width=1\linewidth]{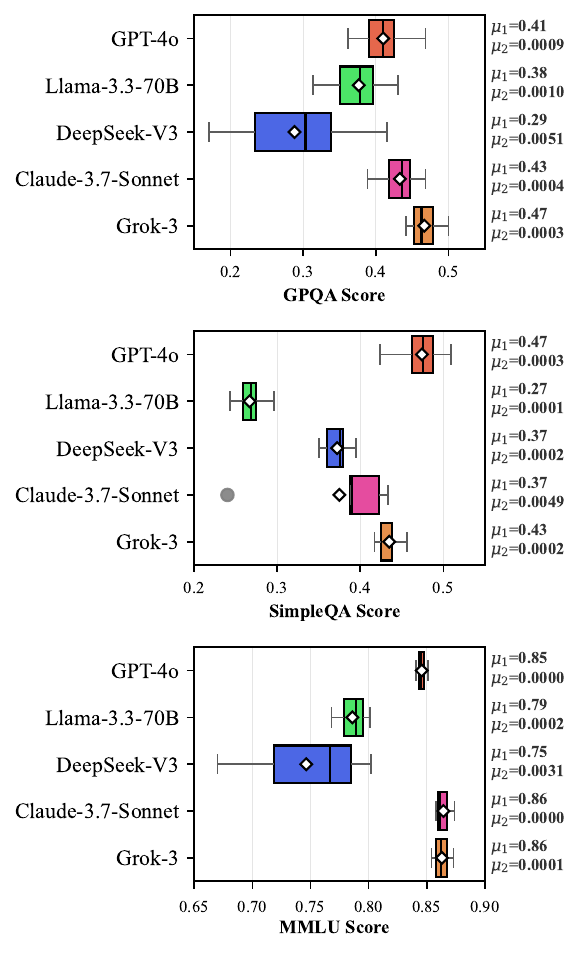}
    \caption{Evaluation of frontier LLMs on multiple \prompts{} following \name{}, estimating the complete prompt sample space. Models vary in both expected value and variance, highlighting the importance of stochastic evaluation.
    }
    \label{fig:all-box-plots}
\end{figure}

A host of recent work has noticed that LLMs are highly sensitive to seemingly arbitrary \emph{prompt perturbations}, throwing into question many of the results reported on popular benchmarks. These perturbations span various dimensions: semantically-equivalent paraphrases of the task instructions~\cite{mizrahi-etal-2024-state}, changes in delimiters or whitespace~\cite{sclar2024quantifying,voronov-etal-2024-mind}, the order of in-context few-shot examples~\cite{lu-etal-2022-fantastically}, among many others~\cite{perlitz-etal-2024-efficient,levy-etal-2024-task,liu-etal-2024-lost}.


While these works observed that LLMs are highly sensitive to prompt perturbations, to the best of our knowledge there is currently no prescriptive recipe for conducting meaningful evaluation which takes this sensitivity into account.
Evidently, many recent evaluation efforts resort to reporting LLM performance against a single arbitrary prompt, while often acknowledging that this practice is flawed~\cite{gu-etal-2024-counterfeit, 10.5555/3692070.3692729}, 
highlighting the need for new evaluation practices.

In this work, we argue that the evaluation of such sensitive LLMs requires \emph{ stochastic evaluation} over the spectrum of perturbations via a method of moments analysis (expected value, variance, etc.). To estimate  moments over the combinatorially large perturbation sample space, we define the notion of \emph{reliable evaluation}, which bounds the probability that a sample of prompt perturbations is representative of the entire sample space. Further, we formulate \name{} -- a simple recipe for estimating the number of samples needed to achieve reliable evaluation per dataset.


Using our recipe, we perform stochastic evaluation of \nummodels{} frontier models, as well as leading open-source models, on \numbenchmarks{} popular benchmarks. Our findings, shown in Figure~\ref{fig:all-box-plots}, reveal the statistical differences between models, highlighting the need for stochastic evaluation. Moreover, we show that the number of resamplings required to reliably estimate model performance varies depending on both the model and the dataset being evaluated.

We hope that our recommendations will be adopted to achieve meaningful and reliable reporting of LLM performance.


\begin{figure*}[h!]
    \centering
    \begin{subfigure}[t]{0.45\textwidth}
        \centering
\includegraphics[width=\linewidth]{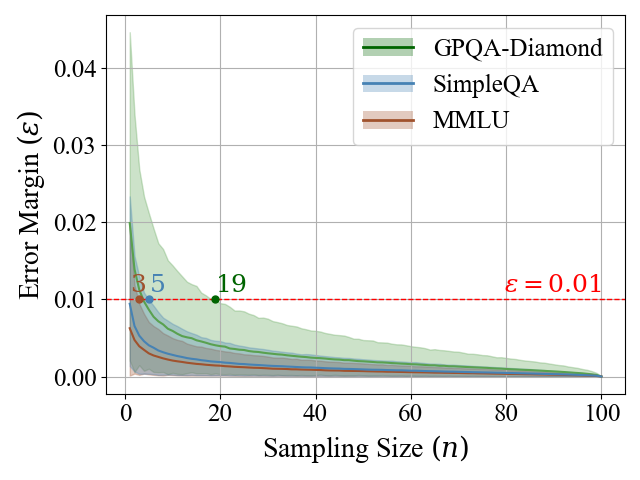}
\caption{\llama{} convergence on different benchmarks.}
\label{fig:conv-llama}
    \end{subfigure}
    \hfill
    \begin{subfigure}[t]{0.45\textwidth}
        \centering
\includegraphics[width=\linewidth]{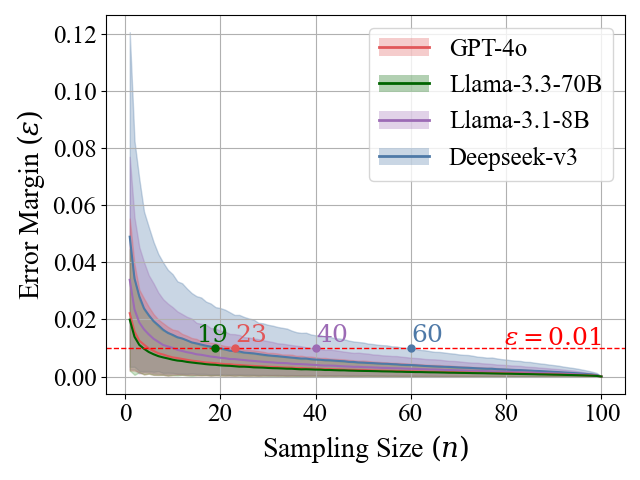}
\caption{Different models convergence on \gpqa{}.}
\label{fig-conv-gpqa}
    \end{subfigure}
\caption{Convergence of the deviation from the true mean accuracy with increasing resampling size. Round markers indicate $n^*$, the min. resamplings as defined in Eq.~\ref{eq-mean}, shown per benchmark in (a) and per model in (b).}
    \label{fig:all-convergence}
\end{figure*}


\section{Stochastic Evaluation of LLMs: Desiderata and Approximation}\label{sec:stochastic}

Here we propose a set of desired metrics for 
LLM evaluation in light of their observed 
sensitivity~(\S\ref{sec:desired}). Since computing these metrics directly is infeasible, we also describe the desired statistical properties of a reliable approximation~(\S\ref{sec:approx}). 
In the following sections, we will operationalize these concepts (§\ref{sec:recipe}), and use this approach to evaluate frontier LLMs (§\ref{sec:eval}).

\subsection{Characterizing LLM Performance Using Distributional Analysis}\label{sec:desired}
We formulate the behavior a model $M$ as a random variable with respect to a deterministic evaluation metric $\varepsilon$:
\begin{equation}
\varepsilon_{M}: S_D \mapsto \mathbb{R}_{+}
\end{equation}
Where $D$ denotes an evaluation dataset (e.g., MMLU), the sample space $S_D$ denotes the space of all meaning-preserving prompt perturbations of $D$ (e.g., different instruction paraphrases, different answer enumerators, addition or removal of whitespace), 
and $\varepsilon_M(s)$ denotes the performance of model $M$ on a single prompt $s \in S_D$ according to metric $\varepsilon$. 
For example, $\varepsilon_M(s) \in [0,1]$ can denote the exact-match accuracy  of Llama ($M$) on a single MMLU instance under prompt $s$.
Using this notation, the limitations of current evaluations are evident -- they report the values of $\varepsilon_M$ on arbitrary samples from $S_D$, while aiming to make claims about the entire sample space $S_D$.

\paragraph{A statistically-meaningful evaluation of LLMs.}
This stochastic formulation of LLM performance gives rise to a \emph{method of moments} analysis of its behavior~\cite{casella2024statistical}. 
In particular, we treat \( s \in S_D \), as i.i.d.  resulting from uniform sampling over \( S_D \). I.e., since we focus on meaning-preserving prompt perturbations, they are considered to be equally likely. We further focus on the first and second moments of \( \varepsilon_M \).

The first moment $\mu_1$ denotes the model's \emph{expected value} over the space of all meaning-preserving prompt perturbations: 

\begin{equation}
\label{eq:mean}
\begin{aligned}
    \mu_1(M, S_D) &= \underset{s \overset{\text{i.i.d.}}{\sim} S_D}{\mathbb{E}}[\varepsilon_M] \\ 
    &=  \sum_{s \in S_D} \varepsilon_M(s) \cdot P(S = s)  \\
    &\overset{\text{\tiny{uniform i.i.d.}}}{=} \frac{1}{\big|S_D\big|} \sum_{s \in S_D} \varepsilon_M(s)
\end{aligned}
\end{equation}

Similarly, the second moment $\mu_2$, i.e., \emph{variance}, is given by:

\begin{equation}
\label{eq:var}
\begin{aligned}
    \mu_2(M, S_D) &= \mathbb{E}_{s \overset{\text{i.i.d.}}{\sim}
 S_D}[{\varepsilon_{M}}^{2}] \\
 &= \mathbb{E}_{s \overset{\text{i.i.d.}}{\sim}
 S_D}[(\varepsilon_M(s) - \mu_1)^{2}]  \\
    &\overset{\text{\tiny{uniform i.i.d.}}}{=} \frac{1}{\big|S_D\big|} \sum_{s \in S_D} \left( \varepsilon_M(s) - \mu_1 \right)^2
\end{aligned}
\end{equation}

This framework allows future work to extend the analysis to additional moments and other distributions beyond uniform i.i.d~\cite{siska-etal-2024-examining}.




\subsection{Reliable Estimation of Distributional Analysis}\label{sec:approx}
Note that explicitly computing the moments in Equations \ref{eq:mean} and \ref{eq:var} is infeasible, as it requires knowing the entire space of meaning-preserving prompt perturbations, which explodes combinatorially (e.g., for all of the permutations of few shot examples) and is even hard to enumerate (e.g., such is the case for the space of all instruction paraphrases). 
Instead, we aim to estimate these moments using a random sample $S' \subset S_D$, relying on the linearity of expectations, as is similarly done in stochastic gradient descent. 

 Below we define dataset-specific requirements to make sure that $S'$ is large enough to enable reliable estimation of the true moments.

 \begin{definition}[Reliable evaluation]
 \label{def:reliable}
 Given an error margin \( \epsilon \) and confidence level \(\delta\), let \( S_D \) be the space of all meaning-preserving prompt perturbations of dataset \( D \), and let \( S' \subset S_D \) be a random subset of size \( n \). Then, we say that $n$ samples yield a \emph{reliable evaluation} if 
 for every moment $\mu_i$ (expected value and variance),
 it holds that:
 \begin{equation}\label{eq:reliable}
\underset{\substack{S' \subset S_D \\|S'| = n}}{\mathbf{P}} \bigg[ \big| \mu_i(M, S') - \mu_i(M, S_D) \big| > \epsilon \bigg] < \delta
 \end{equation}
 \end{definition}

 In other words, an evaluation based on \( n \) resamplings of \( S' \subset S_D \) with $\big|S'\big|=n$ is considered \textit{reliable} if the probability that the empirical momentum of the sample $S'$ deviates from the momentum over the entire distribution by more than $\epsilon$ is bounded by $\delta$. 
In section \ref{sec:recipe} we propose a method for estimating the required \( n \), by constructing a confidence interval around this deviation. 

 We can then perform stochastic evaluation over this reduced resampling space, reporting empirical moments which are expected to yield with high probability a good estimation of the true moments over the entire sample space.

 





\section{\name{}: Recipe for Stochastic Evaluation}

In this section, we present a practical recipe for conducting a reliable stochastic evaluation of LLMs. 

The recipe assumes a scenario aiming to evaluate a set of models \( M_1, \dots, M_k \) on a dataset \( D \), while accounting for LLMs' sensitivity to \prompts. 


\paragraph{Step 1: Specify evaluation parameters \( \bm{\epsilon} \) and \( \bm{\delta} \).}  
Set the acceptable deviation \( \epsilon \) between the empirical value of the \( i \)-th moment over a sample \( S' \subset S_D \) and the corresponding moment over the full distribution \( S_D \), as well as the confidence level \( \delta \) with which this guarantee should hold, as defined in Equation~\ref{eq:reliable}.
In particular, we propose to set $\epsilon = 0.01$ and $\delta = 0.1$, i.e., that evaluation should be considered reliable if it deviates from true distribution by no more than $0.01$ with probability of at least $0.9$. This can critically examine claims of state of the art performance, which typically revolve around a difference of a few performance points between models~\cite{liu2024deepseek}. 

\paragraph{Step 2: Define the sample space of meaning-preserving paraphrases \( \bm{S_D} \).}  
Identify dimensions of meaning-preserving prompt perturbations that may influence model performance -- such as instruction phrasing, output format, or few-shot examples. We recommend leveraging existing work aligned with the task type. For instance, for multiple-choice QA datasets, the framework by~\citet{habba2025dove} can be used to generate the prompt perturbation space \( S_D \). Their approach builds on the Unitxt framework for structured data preparation, which can also be extended to generate prompt perturbations for other task types~\cite{bandel-etal-2024-unitxt}. 
Notably, our proposed method is flexible and not restricted to any predefined set of \prompts, and other paraphrases can be used to construct the sample space $S_D$.


\paragraph{Step 3: Estimate the minimal reliable sample size \( \bm{n^*} \).}
Our goal here is to identify the smallest sample size \( n \) which satisfies the reliability condition in Definition~\ref{def:reliable}. 
This is challenging since it requires computing true moments over the entire distribution. To estimate this, we propose to choose a reference model $\hat{M}$ and compute its empirical moments over large $N$ as proxy for true moments. In the following section, we will show that choosing a relatively cheap model gives empirically good estimates, which hold across models.
For each candidate sample size \( n = 1, 2, \ldots, N \),  compute the set of deviations between the empirical value of the \( i \)-th moment over each subset \( S' \subset S_D \) of size \( n \), and the \( i \)-th moment computed over $N$ samples:
\begin{align}\label{eq:delta-dev}
    \Delta(n) = \Big\{\, 
         \big| \mu_i(M, S') - \mu_i(M, S_D) \big| : 
        |S'| = n 
    \,\Big\}
\end{align}

Next, construct the \( \delta \)-level confidence interval (CI) over \( \Delta(n) \), which filters $\Delta(n)$ to the range between the \( \delta/2 \) and \( 1 - \delta/2 \) percentiles. For instance, if \( \delta = 0.1 \), the corresponding $\mathbf{CI}_{0.1}(\Delta(n))$ includes all values of $\Delta(n)$ which lie between the 5th and 95th percentiles.  
    Then, define \( n^* \) as the smallest \( n \) for which \( \epsilon \) is larger than the maximum of this confidence interval:
    \begin{equation}\label{eq-mean}
        n^* = \min \left\{ n \in [1, N] \;\middle|\; \epsilon \geq \max{\mathbf{CI}_\delta(\Delta(n))} \right\}
    \end{equation}

We note that in some scenarios, such as when the focus is on evaluating a single model or when the variations between models is large, it may be preferable to use a reference dataset instead of a reference model. For example, if we want to evaluate model $M$ on multiple datasets, we can choose a reference dataset $D’$, compute its empirical moments over large $N$ as a proxy for the true moments.

\paragraph{Step 4: Report empirical distribution analysis.}

Finally, sample a subset of perturbations \( S' \subset S_D \) of size \( |S'| = n^* \) uniformly at random. Then, evaluate each model \( M_1, \dots, M_k \) on all prompt variations \( s \in S' \), and report empirical moment analysis. In particular, we recommend reporting box plot showing median and interquartile range of observed performance, as can be seen in Figure~\ref{fig:all-box-plots}.

\label{sec:recipe}

\section{Reliable Stochastic Evaluation of Frontier Models}\label{sec:eval}

In this section, we present a reliable stochastic evaluation of \nummodels{} state-of-the-art LLMs, including both open-source and proprietary models, across \numbenchmarks{} widely used benchmarks.

\subsection{Experimental Setup}

We run \name{} on \mmlu{}~\cite{hendrycks2021measuring}, \gpqa{}~\cite{rein2024gpqa}, and \simple{}~\cite{wei2024measuring}, which are all widely-used English benchmarks. The curation of the \prompts{} space is done by leveraging unitxt~\cite{bandel-etal-2024-unitxt} and Dove~\cite{habba2025dove}.
 We evaluate five LLMs: \llama{}~\cite{grattafiori2024llama}, \deepseek{}~\cite{liu2024deepseek}, \gpt{}~\cite{hurst2024gpt}, \claude{}~\cite{claude37sonnet}, and \grok{}~\cite{grok3}. 
As defined in Section~\ref{sec:stochastic}, we set the following parameters to estimate a reliable evaluation
$
\epsilon=0.01, \: \delta = 0.1, \: N = 100
$,
with \llama{} serving as the reference model $\hat{M}$ for estimating $n^*$.
See additional implementation details in the Appendix.

\subsection{Results}

\paragraph{Frontier models are sensitive to meaning-preserving prompt perturbations, underscoring the need for stochastic evaluation.}
Figure~\ref{fig:all-box-plots} shows that across all three evaluated benchmarks, model performance varies across different prompt resamplings. 
This highlights the importance of stochastic evaluation, i.e., reporting statistical measures over the distribution of scores rather than relying on single prompts. 
As shown by the overlapping boxplots in Figure~\ref{fig:all-box-plots}, there is \emph{often no definitive winner} -- any meaning-preserving prompt could be cherry-picked to suggest a particular model ranking. 

\paragraph{The number of resamplings required for reliable evaluation depends both on the dataset and on the model.}
In Figure~\ref{fig:conv-llama}, we show that the convergence behavior of \llama{}'s estimation depends on the benchmark.
Moreover, in Figure~\ref{fig-conv-gpqa}, we observe that different models exhibit different convergence rates on the same dataset, suggesting that reliable evaluation is determined by both the model and the dataset.


\paragraph{Llama-3.1-8B can guide the number of resamplings needed for reliable evaluation of \llama{}.}
While \llama{} substantially outperforms the smaller Llama-3.1-8B, Figure~\ref{fig-conv-gpqa} shows that the smaller model provides a valid upper bound on convergence behavior. This suggests that smaller models can serve as effective proxies for estimating the number of prompt resamplings required for reliable stochastic evaluation of larger models. This is shown also for the \gpqa{} and \simple{} (Figure~\ref{fig:all-big-vs-small} in Appendix). 

\section{Related Work}
Most related to our work, \citet{polo2024efficient} proposed a method for multi-prompt evaluation, hinging on a binary Bernoulli distribution, limiting its applicability to text generation,  and revolving around the selection of representative evaluation examples. In contrast, we  find a minimal representative random subspace, are agnostic to the type of perturbations, and do not make any assumption about the scoring function. Other works highlight the importance of multi-prompt evaluations , albeit without prescriptive guidelines ~\cite{voronov-etal-2024-mind, tam-etal-2024-speak, zhuo-etal-2024-prosa, hida2024social}.

\section{Conclusion}
We propose to estimate model performance over prompt variations using moment analysis and show how to compute how many samples are needed for reliable results. 

Our proposed method is designed to accommodate any computational budget, with an inherent trade-off between budget, error margin, and confidence. The practical question becomes: \textit{``Given a specific compute budget, what is the most reliable evaluation achievable?''} In our framework, the compute budget sets the maximum feasible $N$ and constrains $n^*$. If for a given error margin $\epsilon$ and confidence level $\delta$ the number of samples $n*$ exceeds a given budget, it is still possible run the evaluation with $n < n^*$ resamplings, accepting a larger margin of error or lower confidence as a result. Thus, even with limited compute resources, our method provides guidance on how to maximize evaluation reliability within those constraints.

Finally, by evaluating frontier models across benchmarks, we find that sensitivity varies widely, underscoring the need for more robust evaluation practices.




\section*{Limitations}

We identify several limitations of this work that future research may address.

First, \name{} requires running a reference model $\hat{M}$ over a large number of resamplings $N$. While this is performed only once, it can be computationally expensive—especially in LLM-based evaluation settings where the reference model also serves as a judge and is costly to query.

Second, there are two additional factors that may influence the required resampling size, which we did not directly investigate. Future work may explore: (1) the effect of dataset size on the number of resamplings needed, and (2) the impact of the model's decoding strategy, which is known to affect evaluation outcomes~\cite{song-etal-2025-good}. For the latter, we provide an initial comparison in Figure~\ref{fig:greedy}, showing results for \gpt{} using greedy decoding versus sampling with a default temperature. However, further experimentation is needed to better understand these effects.

\section*{Acknowledgments}This work was partially supported by research grant no. 7256
from the Israeli Ministry of Science and Technology. We thank Dr. Arie Cattan and Dr. Ori Shapira for the helpful discussions and advice on this project.

\bibliography{anthology,custom}

\appendix

\section{Appendix}
\label{sec:appendix}
\subsection{Benchmarks and Prompt Perturbations}\label{app:bench}

We provide additional details about the benchmarks used in our evaluation with \name{}.

\paragraph{Prompt Perturbation Dimensions.}
For each benchmark, we define task-specific dimensions of prompt perturbations over which we resample.

For \mmlu{} and \gpqa{} (Multiple-Choice QA), we follow the resampling strategy from~\cite{habba2025dove}, varying along four dimensions:  
(1) instruction paraphrasing,  
(2) answer choice order,  
(3) answer choice enumerator (e.g., letters, numbers, Roman numerals), and  
(4) choice separators (e.g., whitespace, tab, newline) between the answers.

For \simple{} (Open-Ended QA), we vary:  
(1) instruction phrasing (e.g., ``Answer the following question''),  
(2) which examples are selected for evaluation,  
(3) the selection and ordering of few-shot demonstrations, and  
(4) whether prompts include `Question:' and `Answer:' markers.

\paragraph{Number of Examples Per Benchmark.}
For \gpqa{}, we evaluate the full dataset, with 198 examples per resampling.  
For \mmlu{}, we sample 100 examples from each subcategory, resulting in 5,700 total examples (from the 14K test split), reused across all resamplings.  
For \simple{}, which includes variation in the evaluation examples themselves, we randomly select 1K examples (from 4K) per resampling, ensuring full coverage over multiple runs.

\paragraph{Prompting Technique.}
We use 5-shot prompting for all benchmarks during evaluation.

\FloatBarrier

\begin{table*}[b]
\resizebox{\textwidth}{!}{%
\begin{tabular}{@{}lllll@{}}
\toprule
\textbf{Model} & \textbf{Version} & \textbf{Decoding Temp.} & \textbf{Inference Platform/API} & \textbf{Total Cost (\$)} \\ \midrule
\gpt & gpt-4o-2024-08-06 & 1.0 & OpenAI & 100 \\
\llama & Llama-3.3-70B-Instruct-Turbo & 0.0 & Together AI & 420 \\
\deepseek & DeepSeek-V3 & 0.3 & Together AI & 60 \\
\grok & grok-3 & 0.1 & XAI & 60 \\
\claude & claude-3-7-sonnet-20250219 & 0.0 & Anthropic & 60  \\
Llama-3.1-8B & meta-llama/Llama-3.1-8B-Instruct & 0.0 & vLLM on local a6000 (1) & $N/A$ \\
 \bottomrule
\end{tabular}%
}
\caption{Model inference configurations.}
\label{tab:models}
\end{table*}


\begin{figure*}[b]
    \centering
    \begin{subfigure}[t]{0.45\textwidth}
        \centering
\includegraphics[width=\linewidth]{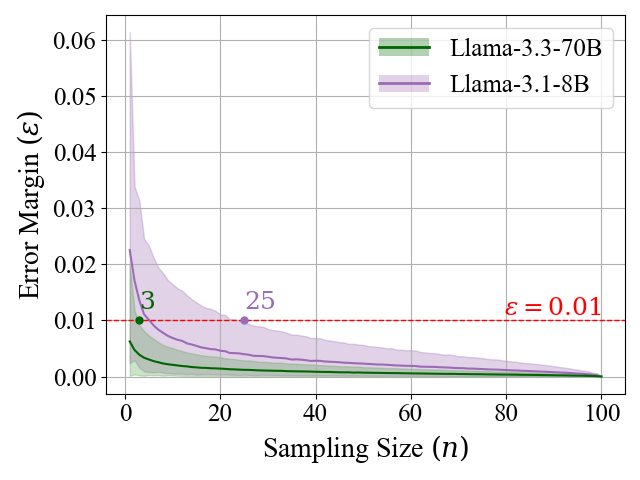}
\caption{\mmlu}
    \end{subfigure}
    \hfill
    \begin{subfigure}[t]{0.45\textwidth}
        \centering
\includegraphics[width=\linewidth]{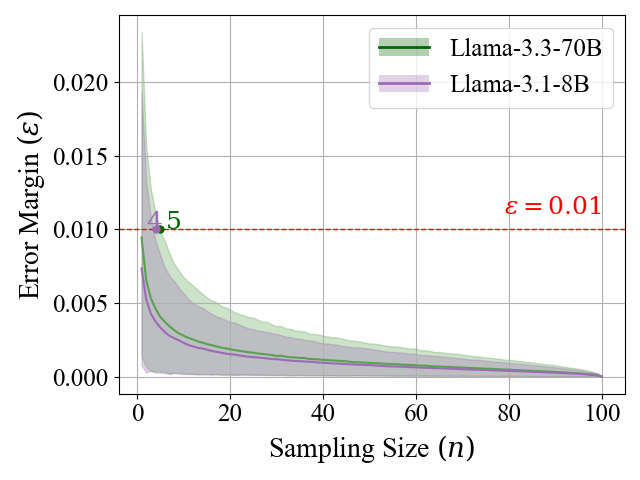}
\caption{\simple{}}
    \end{subfigure}
    \caption{Error convergence of \llama{} vs Llama-3.1-8B.}
    \label{fig:all-big-vs-small}
\end{figure*}
\begin{figure*}[hb!]
\includegraphics[width=0.45\linewidth]{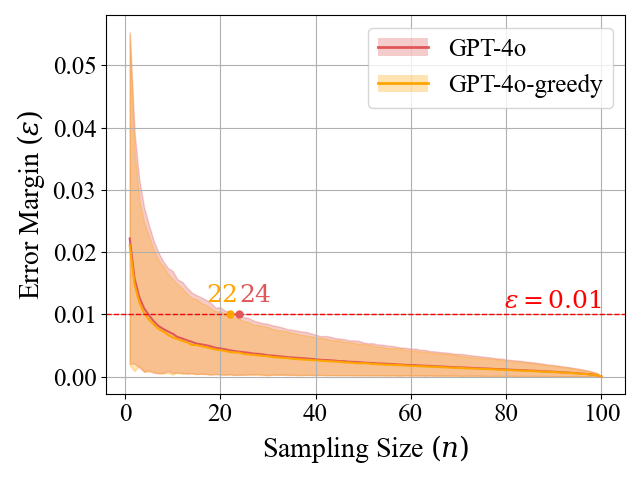}
    \caption{
        \gpt{}'s error convergence on \gpqa{},\\
        greedy decoding versus default temperature sampling (temp=1).
    }
    \label{fig:greedy}
\end{figure*}


\subsection{Evaluation Setup}\label{sec:app-details}

\paragraph{LLM-as-a-Judge for \simple{}.}
To evaluate \simple{}, we use an LLM-as-a-judge setup to determine alignment between predictions and gold answers.  
We adopt the judging prompt from the official \simple{} repository.\footnote{\url{https://github.com/openai/simple-evals}}  
Our judge model is Atla Selene Mini~\cite{alexandru2025atla}, which currently ranks highest among open-source models on the Judge Arena Leaderboard.\footnote{\url{https://huggingface.co/spaces/AtlaAI/judge-arena}}

\paragraph{Model Decoding Temperatures.}
To match typical usage, we adopt model-specific decoding temperatures aligned with standard evaluation practices, informed by official documentation and community reports.

\end{document}